\documentclass[final]{l4dc2024}

\title[PlanNetX]{PlanNetX: Learning an Efficient Neural Network Planner from MPC for Longitudinal Control}
\usepackage{times}
\usepackage{booktabs}
\usepackage{amsmath, amssymb, amsfonts, mathtools}
\usepackage{xcolor}
\usepackage{makecell}
\usepackage{graphicx}
\usepackage[short]{optidef}
\usepackage{wrapfig}

\usepackage{amsmath,amsfonts,bm}

\def\figref#1{figure~\ref{#1}}

\def\1{\bm{1}}
\newcommand{\train}{\mathcal{D}}
\newcommand{\valid}{\mathcal{D_{\mathrm{valid}}}}
\newcommand{\test}{\mathcal{D_{\mathrm{test}}}}

\DeclareMathAlphabet{\mathsfit}{\encodingdefault}{\sfdefault}{m}{sl}
\SetMathAlphabet{\mathsfit}{bold}{\encodingdefault}{\sfdefault}{bx}{n}

\newcommand{\E}{\mathbb{E}}

\newcommand{\R}{\mathbb{R}}

\coltauthor{%
 \Name{Jasper Hoffmann} \Email{hoffmaja@informatik.uni-freiburg.de} \\
 \Name{Diego {Fernandez Clausen}} \Email{fernandi@informatik.uni-freiburg.de} \\
 \Name{Julien Brosseit} \Email{brosseit@informatik.uni-freiburg.de} \\
 \addr Neurorobotics Lab, University of Freiburg, Germany.
 \AND
 \Name{Julian Bernhard} \Email{julian.bernhard@bmw.de} \\
 \Name{Klemens Esterle} \Email{klemens.esterle@bmw.de} \\
 \Name{Moritz Werling} \Email{Moritz.Werling@bmw.de} \\
 \Name{Michael Karg} \Email{Michael.Karg@bmw.de} \\
 \addr BMWGroup, Unterschleissheim, Germany.
 \AND
 \AND
 \Name{Joschka Bödecker} \Email{jboedeck@informatik.uni-freiburg.de} \\
 \addr Neurorobotics Lab, University of Freiburg, Germany.
}

\begin{document}

\maketitle

\begin{abstract}%
Model predictive control is a powerful, optimization-based approach for controlling dynamical systems.
However, the computational complexity of online optimization can be problematic on embedded devices.
Especially, when we need to guarantee a fixed control frequency.
Thus, previous work proposed to reduce the computational burden using imitation learning approximating the MPC policy by a neural network.
In this work, we instead learn the whole planned trajectory of the MPC.
We introduce a combination of a novel neural network architecture PlanNetX and a simple loss function based on the state trajectory that leverages the parameterized optimal control structure of the MPC.
We validate our approach in the context of autonomous driving by learning a longitudinal planner and benchmarking it extensively in the CommonRoad simulator using synthetic scenarios and scenarios derived from real data.
Our experimental results show that we can learn the open-loop MPC trajectory with high accuracy while improving the closed-loop performance of the learned control policy over other baselines like behavior cloning.
\end{abstract}

\begin{keywords}%
 imitation learning, model predictive control, autonomous driving%
\end{keywords}

\section{Introduction}
Motion planning for autonomous driving is challenging due to the high uncertainty of the surrounding traffic requiring constant re-planning to adjust the plan to newly obtained information.
Model predictive control (MPC) is an optimization-based approach that relies on rapidly resolving an optimal control problem (OCP) numerically, planning a trajectory at each step.
However, online optimization can be a limiting computational factor if a control needs to be provided in a fixed time interval and especially if a lot of different candidate trajectories need to be planned in parallel.

To reduce the computational cost of MPC, a lot of effort was put into computing the MPC solutions offline and storing them in an appropriate representation to allow fast online evaluation during deployment.
One example of this is \emph{explicit MPC}.
In the case of linear MPC, the MPC policy, defined as the first control of the MPC solution, is a piece-wise linear function that can be stored as a look-up table \cite{Bemporad1999}.
However, for non-linear dynamics or constraints, the exact representation of an MPC policy is challenging and often done only approximately \cite{Johansen2003}.
Another prominent example is imitation learning (IL).
Here, we try to approximate the MPC policy by using function approximators like neural networks.
Previous work focused on improving the safety aspects \cite{mamedovSafeImitationLearning2022} of the learned policy or replacing the loss function of standard behavior cloning (BC) \cite{cariusMPCNetFirstPrinciples2020, ghezziImitationLearningNonlinear2023}, which tries to minimize the squared distance to the first control of the MPC solution, using objectives that are aware of the underlying OCP structure of the MPC.

In this work, instead of just learning the first control we want to learn the whole planned trajectory of the MPC solution.
This is useful in the field of autonomous driving, where it is common to separate the tasks of motion planning from the task of low-level vehicle control to simplify the control problem.
Additionally, we find that learning the whole planned trajectory allows us to define a simple loss formulation that uses the underlying OCP structure during training being potentially aware of the network's approximation errors.
Furthermore, learning the trajectory could be used to warm start the MPC solver for significantly faster convergence and better safety guarantees.
The goal of this work is to general improve the planning time in comparison to numerically solving the OCP.
Thus, we deploy common neural network compression techniques like pruning and quantization to reduce the inference time.

The contributions of this work are threefold:
    \textbf{(1)} We propose an imitation learning framework called \texttt{PlanNetX} specifically for learning the planned trajectory of an MPC planner, consisting of a simple loss function and a specialized neural network architecture.
    \textbf{(2)} We validate the approach by learning a longitudinal planner for autonomous driving and extensively testing it on a handcrafted CommonRoad, \cite{althoffCommonRoadComposableBenchmarks2017}, benchmark set consisting of real and synthetic driving data.
    \textbf{(3)} We analyze whether common neural network compression techniques like pruning and quantization allow us to reduce the inference time even further.

\subsection{Related Work}
Recent autonomous driving systems incorporate more and more learning-based control \cite{kiranDeepReinforcementLearning2022b}, by learning control policies via IL or reinforcement learning (RL) either from simulations or human driving data.
For autonomous driving, RL was used to warm-start the MPC solver in order to
mitigate the problem of the solver getting stuck in local minima \cite{hartLaneMergingUsingPolicybased2019}.

In the context of IL from an MPC policy, various aspects were considered:
(1) \emph{Safety and Stability}: 
\cite{mamedovSafeImitationLearning2022} use a safety filter formulation similar to \cite{wabersichPredictiveSafetyFilter2021} to guarantee the safety of the final control policy.
Using the concept of control-barrier functions,  \cite{cosnerEndtoEndImitationLearning2022} show that under certain conditions the safety guarantees of a robust MPC expert can be transferred to the learned policy.
(2) \emph{Verification}:
\cite{schwanStabilityVerificationNeural2023} provide an upper bound on the approximation error of the learned policy to the original MPC policy by solving a mixed-integer formulation, whereas in \cite{hertneckLearningApproximateModel2018, kargProbabilisticPerformanceValidation2021} a probabilistic bound is given.
(3) \emph{Improved Loss Function}:
Further research tries to replace the surrogate loss of standard behavior cloning, which is often the squared distance, with an objective that takes into account the underlying OCP problem:
In \cite{cariusMPCNetFirstPrinciples2020, reskeImitationLearningMPC2021} a loss based on the control Hamiltonian is introduced, whereas in \cite{ghezziImitationLearningNonlinear2023} a Q-loss is introduced by calculating the value of a fixed first control with an MPC formulation.

IL from trajectory optimization was considered by \cite{mordatchCombiningBenefitsFunction2014, levineEndtoendTrainingDeep2016} where an alternating direction method of multipliers (ADMM) method jointly optimizes a task and policy reconstruction cost.
Similarly, in \cite{levineGuidedPolicySearch2013, levineVariationalPolicySearch2013} an iterative linear quadratic regulator (iLQR) \cite{li2004iterative} is used for guiding an RL policy to low costs regions.
A related learning formulation to the one presented in this work is \cite{drgonaDeepLearningExplicit2021, drgonaLearningConstrainedParametric2024} where the learning objective is derived from an OCP by differentiating through the dynamics to solve a tracking formulation.
Our approach differs in that we first solve the OCP via numerical optimization and then use the solution as a tracking signal.

In general, apart from \cite{drgonaDeepLearningExplicit2021, drgonaLearningConstrainedParametric2024, vaupelAcceleratingNonlinearModel2020} most approaches focus on learning the control policy but not the planned trajectory.

\section{Preliminaries}

In this section, we give a quick introduction to MPC and IL.

\subsection{Model Predictive Control}
In the following, we introduce the notion of a parameterized discrete-time OCP.
At each time step, an MPC planner observes a current state $\bar{x}_0 \in \R^{n_x}$ to generate an optimal control and state trajectory $U \in \R^{n_u \times N}$ and $X \in \R^{n_x \times {N+1}}$ by solving the OCP
\begin{mini!}[2]
    {X, U}{T(x_N) + \sum_{k=0}^{N-1} L_k(x_k, u_k) }{\label{ocp: discrete-time OCP}}{\label{cf: discrete-time OCP}}
    \addConstraint{x_{k+1}}{= f(x_k, u_k),}{\quad k=0,\dots, N-1  \label{cns: system dynamic}}
    \addConstraint{h(x_k, u_k, p_k)}{\leq 0,}{\quad k=0,\dots, N-1 \label{cns: h<=0}}
    \addConstraint{x_0}{= \bar{x}_0, \label{cns: initial state constraint}}
\end{mini!}
with a planning horizon length $N$, stage costs $L_k$, terminal costs $T$, and known, potentially simplified, system dynamics $f$.
Furthermore, in \eqref{cns: h<=0} we assume that we can parameterize the inequality constraints $h$ at each planning step with additional parameters $P \in \R^{n_p \times N}$.
For the rest of the work, we denote $X^\star \in \R^{n_x \times {N+1}}$ and $U^\star \in \R^{n_u \times N}$ as the optimal solution found by the MPC when solving the above OCP.

\subsection{Imitation Learning}

With IL we try to approximate the behavior of an expert by using a function approximator like a neural network.
In the context of this work, the MPC planner is the expert.
We differentiate between learning the \emph{control policy} or control law $\bar{x}_0 \mapsto u_0^\star$ of the MPC, which is already enough to control the system,
and learning the \emph{planning policy} of the MPC planner, where we want to approximate the mapping $\bar{x}_0 \mapsto \left( X^\star, U^\star \right)$.
Note, that the second formulation contains the first one as a sub-problem.
In more mathematical terms, we want to approximate either the MPC control or plan by the following parameterized policies
\begin{align}
    \pi_\theta^\textrm{policy}: \R^{n_x} \times \R^{n_p \times N} &\to \R^{n_u}, & \pi_\theta^\textrm{plan}: \R^{n_x} \times \R^{n_p \times N} &\to \R^{n_x \times (N + 1)} \times \R^{n_u \times N}, \label{eq:formulation_mapping} \\
    \bar{x}_0, P &\mapsto \hat{u}_0^\theta & \bar{x}_0, P &\mapsto \hat{X}^\theta \times \hat{U}^\theta  \nonumber
\end{align}
where $\theta$ denotes the parameter vector of the policy which could be for example the weights of a neural network.
For the planning policy $\pi_\theta^\textrm{plan}$, we assume that $\hat{x}_0^\theta = \bar{x}_0$.
The goal of both learning formulations is to minimize the following two objectives:
\begin{align}
    \mathcal{L}^\textrm{policy} (\theta) &= \E_{\bar{x}_0, P \sim \mathcal{D}} \left[ \ell \left(\hat{u}_0^\theta,\; u_0^\star \right) \right],
    &
    \mathcal{L}^\textrm{plan} (\theta) &= \E_{\bar{x}_0, P \sim \mathcal{D}} \left[ \ell \left(\hat{U}^\theta,\;\hat{X}^\theta,U^\star,\;X^\star \right) \right], \label{eq:formulation_loss}
\end{align}
where $\train$ is a given state and parameter distribution over $ \R^{n_x} \times \R^{n_p \times (N + 1)}$ whereas $\ell$ is an arbitrary pointwise loss functions.

\paragraph{Behavior Cloning:}
For learning just the control policy, the simplest formulation is BC which does not use any information about the underlying OCP.
Instead, here we minimize the distance of our predicted control $\hat{u}_0^\theta$ to the optimal control $u^\star_0$ quadratically
\begin{align}
    \mathcal{L}^\textrm{BC}(\theta) \coloneqq \E_{\bar{x}_0, P \sim \mathcal{D}} \left[ \Vert \hat{u}_0^\theta   - u^\star_0 \Vert_2^2 \right].
\end{align}
Recent work tries to replace the quadratic distance with a distance measure that considers the underlying OCP structure as in \cite{ghezziImitationLearningNonlinear2023, cariusMPCNetFirstPrinciples2020, reskeImitationLearningMPC2021}.

\section{Learning an MPC-based Planner}

In the following, we want to discuss how we can use the knowledge of the experts underlying OCP structure to learn the planned trajectory.
We first discuss different loss formulations and then continue with the \texttt{PlanNetX} architecture tailored to learning the MPC planning policy $\pi^\textrm{plan}$ of \eqref{eq:formulation_mapping}.

\subsection{Loss Formulation}

Given the initial state $\bar{x}_0$ and the dynamics model $f$, predicting the controls $\hat{U}^\theta$ is already enough to also generate the according state trajectory $\hat{X}^\theta$.
One idea would be to minimize the distance $\hat{U}^\theta$ between $U^\star$ as done in \cite{vaupelAcceleratingNonlinearModel2020}.
For this, we define the \emph{control trajectory loss} 
\begin{equation}\label{eq:uloss}
    \mathcal{L}^u (\theta) \coloneqq \E_{\bar{x}_0,P \sim \train} \left[ \frac{1}{N} \sum_{k=0}^{N - 1} \gamma^k \left\lVert \hat{u}_k^\theta - u_k^\star \right\rVert_W^2  \right],
\end{equation}
where $\gamma$ is a discount factor, with $0 < \gamma \leq 1$ allowing to put more weights on early controls in the loss, and  $W$ is a positive-definite weight matrix, defining a norm via $\lVert x \rVert_W^2 = x^T W x$.
However, the loss function defined in \eqref{eq:uloss} does not consider any approximation errors of the neural network.
Thus, if there is a mismatch early in the predicted trajectory, there is no incentive for the network to correct it in the later parts of the predicted trajectory.

Instead of minimizing the distance between $\hat{U}^\theta$ and $U^\star$ in control space, we instead minimize the distance in the state space, thus between the predicted state trajectory $\hat{X}^\theta$ and $X^\star$.
We define the \emph{state trajectory loss} with
\begin{align}\label{eq:xloss}
    \mathcal{L}^{x} (\theta) \coloneqq \E_{\bar{x}_0, P \sim \train} \left[ \frac{1}{N} \sum_{k=1}^{N} \gamma^k \left\lVert \hat{x}_k^\theta - x_k^\star \right\rVert_W^2  \right].
\end{align}
If there are specific constraints and costs regarding the control trajectory, one could also potentially combine the two loss formulations \eqref{eq:uloss} and \eqref{eq:xloss}.
Note, that using the above loss formulations we are implicitly able to learn the underlying expert cost function and the respective constraints.

\subsection{PlanNetX}

\begin{figure}[t]
    \centering
    \def\svgwidth{1.0\textwidth}
    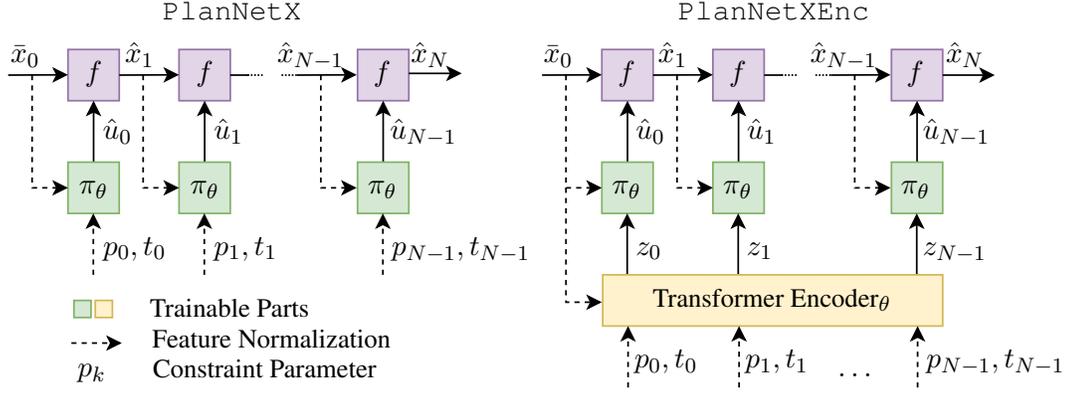
    \caption{The two proposed architectures \texttt{PlanNetX} and \texttt{PlanNetXEnc}.
    The \texttt{PlanNetXEnc} includes an additional Transformer encoder layer that contains potentially the information of the parameters of all different time steps.
   }
    \label{fig:plannet}
\end{figure}

The motivation for the PlanNetX architecture comes from taking a look at the original OCP formulation of \eqref{ocp: discrete-time OCP} that we solve for MPC inference.
The idea is to replace the free decision variables $X, U$ with predictions of a neural network policy $\pi$, by starting a roll-out from the initial state $\bar{x}_0$ and iteratively applying our learned policy $\pi_\theta$ and the dynamics $f$.
We then apply the state trajectory loss on the obtained roll-out resulting in the following formulation:

\begin{subequations}\label{cf: plannet}
\begin{gather}
\min_\theta\quad \E_{\bar{x}_0, P \sim \train} \left[ \frac{1}{N} \sum_{k=1}^{N} \gamma^k \left\lVert \hat{x}_k^\theta - x_k^\star \right\rVert_W^2  \right] \\
\hat{x}_{k+1} = f(\hat{x}_k, \hat{u}_k),\;
\hat{u}_k = \pi_\theta (\hat{x}_k, z_k),\; 
z_k = \psi_k^\theta (P, \; T), 
\hat{x}_0 = \bar{x}_0.
\end{gather}
\end{subequations}

The policy $\pi^\theta$ takes an information vector $z_k$ containing the necessary information of the inequality constraint parameter $P$.
The information vector is derived from an additional constraint network $\psi^\theta_k$ that bundles all the information of the inequality constraint parameter $P$.
We additionally introduce a time vector $T=(0, t_d, 2 t_d, \dotsc, (N-1) t_d)^T \in \R^N$ to give the policy the information, for which time step it predicts a control,
where $t_d$ denotes the discretization time of the MPC planner.
Note, that optimizing \eqref{cf: plannet} requires backpropagating through the dynamics function $f$.


We derive two neural network architectures, see \figref{fig:plannet}.
The \texttt{PlanNetX} architecture does not use an additional parameter network $\psi_\theta$.
Instead, we directly give the parameter $p_k$ and time $t_k$ of stage $k$ as an input.
Note, that in some scenarios this is not enough to predict the control $\hat{u}_k$ correctly, and the information of the full parameter matrix $P$ consisting out of $p_1, \dotsc, p_{N-1}$ is required.
Thus, the second architecture \texttt{PlanNetXEnc} adds a Transformer encoder as described in \cite{vaswaniAttentionAllYou2017a} to compress the whole parameter matrix $P$ and the time vector $T$ into a latent vector $Z \in \R^{n_z \times N}$, providing the policy $\pi_\theta$ with the required information.
For the Transformer each parameter and time tuple $p_k$ can be considered as one token, which is first embedded with a linear layer like in \cite{dosovitskiyImageWorth16x162021} and then processed by the self-attention layer.
The time information is added by a learned positional encoding, as often used in Transformer models \cite{devlinBERTPretrainingDeep2019a}.
We further normalize the input features by min-max normalization where the minimum and maximum values are derived from the values that would be obtained when solving the OCP numerically.


%
%

\section{Experiments}

In the following, we present the empirical results of this work.
After introducing the longitudinal MPC Planner, we continue with our benchmarking setup, and then present and discuss the results in terms of open-loop planning performance, closed-loop performance, and worst-case inference time.

\subsection{Longitudinal Planner}


The following MPC formulation is adopted from \cite{gutjahrLateralVehicleTrajectory2017} and \cite{pekFailSafeMotionPlanning2021}.
We define the longitudinal state of the ego car with $x = [s, v, a, j]^T \in \R^4$, where $s$ is the longitudinal position of the front of the car, $v$ is the velocity, $a$ is the acceleration, and $j$ is the jerk.
The input of the system is the snap given by $u(t) = \ddot{a}(t) \in \R$ which is the derivative of the jerk resulting in the linear time-invariant system:
\begin{equation}\label{eq:long system}
\frac{d^4}{dt^4} s(t) = u(t).
\end{equation}
Integrating the dynamics with a time discretization $t_\textrm{d}$ and assuming piecewise constant inputs we get the discrete-time dynamics system
\begin{equation}
    x_{k+1} = f(x_k, u_k) = A_d x_k + B_d u_k.
\end{equation}

The cost function $L$ is a mixture of quadratic costs penalizing high accelerations, jerk, and controls with weights $w_a, w_j, w_u > 0$ for comfort and encourages the progress made in the longitudinal direction with a linear weight $w_s > 0$, resulting in the following loss
\begin{equation}\label{eq:long costs}
   L_k(x, u) = \gamma^k \left(w_a a^2 + w_j j^2 + w_u u^2 - w_s s \right).
\end{equation}
An additional discount factor $\gamma$, can be used to favor cost minimization of short-term over long-term costs.

\begin{figure}[t]
    \centering
    \includegraphics[width=.95\textwidth]{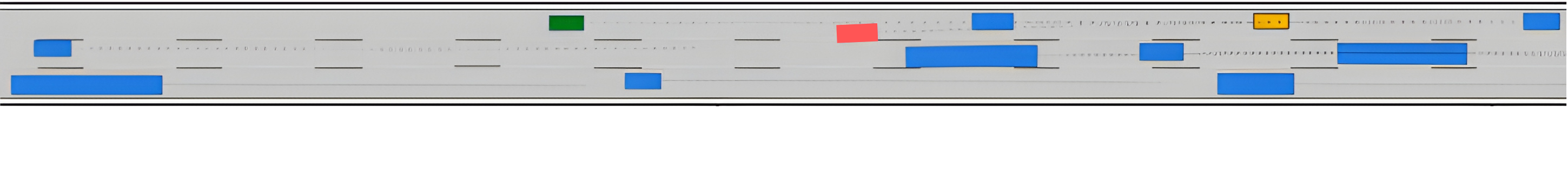}
    \caption{
    Exemplary cut-in scenario of the CommonRoad benchmark created from recordings of the HighD dataset.
    The green car is the ego car, the red car is performing a cut-in in front of the ego car and the yellow region is the goal position the ego car needs to reach.
    The surrounding traffic is depicted in blue.
    }
    \label{fig:CommonRoad}
\end{figure}

To guarantee the feasibility of the planned trajectory, we assume the following box constraints for the velocity $v_\textrm{min} \leq v_k \leq v_\textrm{max}$, acceleration $a_\textrm{min} \leq a_k \leq a_\textrm{max}$ and jerk $j_\textrm{min} \leq j_k \leq j_\textrm{max}$ for $k=0, \dotsc, N$.
We further added a terminal acceleration box constraint, $0 \leq a_N \leq 0$, such that the final acceleration is zero.
We describe the rear position of the \emph{lead vehicle} that is in front of us by $\bar{x}^\textrm{lead}_0 = [s^\textrm{lead}, v^\textrm{lead}, a^\textrm{lead}]^T$ and use a simple forward prediction $\hat{X}^\textrm{lead} \in \R^{n_{x, \textrm{lead}}}$ that assumes that the lead car keeps its acceleration for $t_{\textrm{acc}}$ seconds and then has an acceleration of zero.
With the forward prediction of the lead vehicle $\hat{X}^\textrm{lead}$, a minimum distance $d_\textrm{min}$ and a reaction time $t_\textrm{brake}$, the safety distance to the leading vehicle is assured by the constraint
\begin{equation}\label{eq:cons safety distance}
    h_k^\textrm{dist} (x_k, \hat{x}_k^\textrm{lead}) \coloneqq
        \max \left(
        \frac{v_k^2 - (\hat{v}_k^\textrm{lead})^2}{2 | a_{\textrm{max}}|}
        + v_k \; t_\textrm{brake},\;
        d_\textrm{min} \right) - \left(\hat{s}_k^\textrm{lead} - s_k \right).
\end{equation}
Additionally, to respect speed limits we use the following constraint
\begin{equation}\label{eq:cons speed limit}
    h_k^v (x_k, v^\textrm{max}) \coloneqq v_k - v^\textrm{max} (s_k), \quad \text{with} \; v^\textrm{max} (s_k) \coloneqq 
  \begin{cases}
    v_\textrm{max, 1}     & \quad \text{if } s_k < s_\textrm{change}\\
    v_\textrm{max, 2}     & \quad \text{else}
  \end{cases}
\end{equation}
where $v^\textrm{max}$ is a position-based function encoding the current speed limit, with $v_\textrm{max, 1}$ being the current speed limit, $v_\textrm{max, 2}$ the next speed limit, and $s_\textrm{change}$ the position where the speed limit changes.
Note, that in the context of \eqref{cf: discrete-time OCP} the parameter $p_k$ is given by the forward prediction $\hat{x}^\textrm{lead}_k$ and the constant speed limit parameters $v_{\textrm{max},1}$, $v_{\textrm{max}, 2}$, $s_\textrm{change}$.
Additionally, we introduced a quadratic slack variable $\zeta_\textrm{dist}$ to loosen the distance constraint of \eqref{eq:cons safety distance} with a weight $w_{\zeta, \textrm{dist}} > 0$ and a slack variable $\zeta_{a_N}$ for the terminal acceleration constraint with a weight $w_{\zeta, \textrm{a}_N} > 0$.

\subsection{CommonRoad Benchmark}

\begin{figure}[t]
    \centering
    \includegraphics[width=1.0\textwidth]{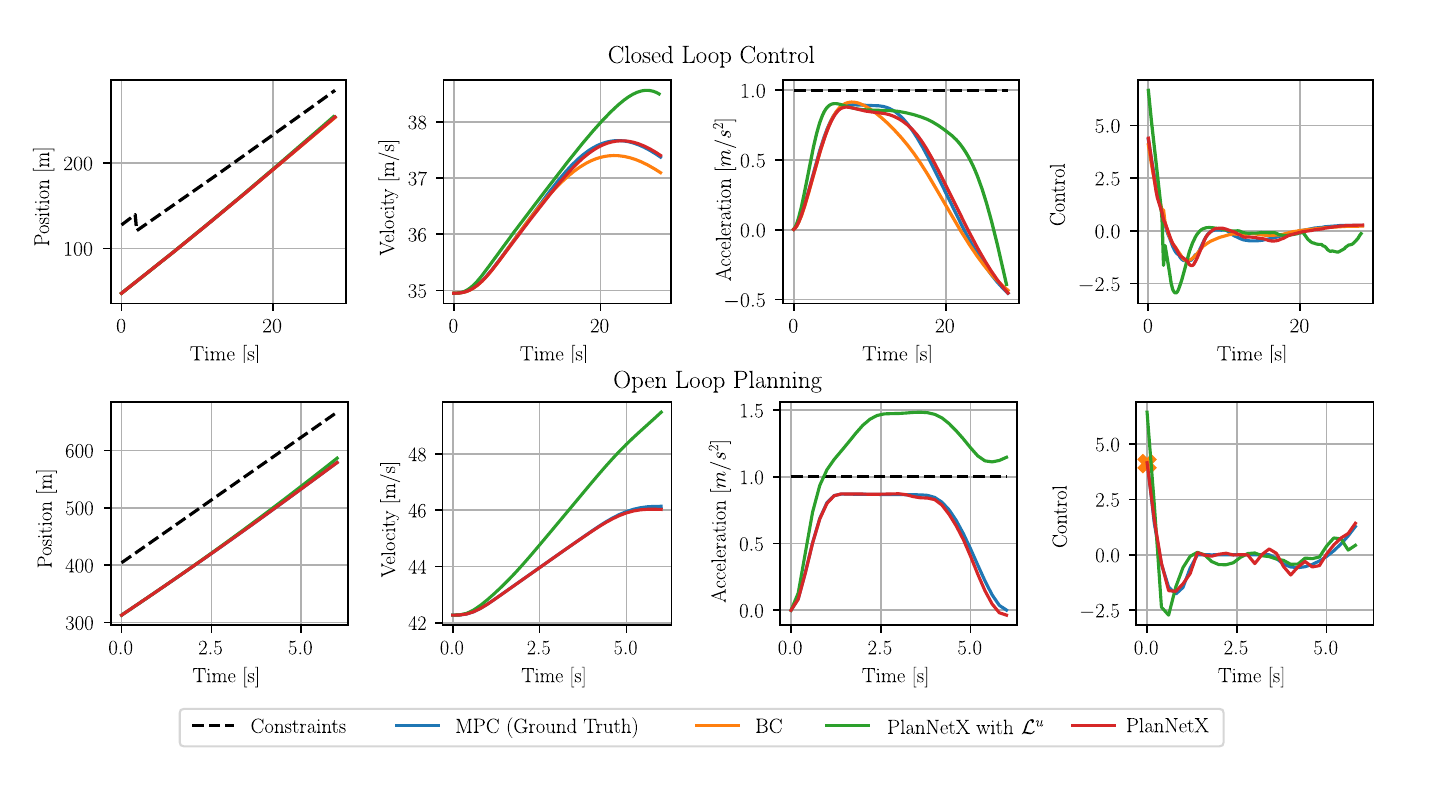}
    \caption{The closed-loop control plots show a trajectory generated in the CommonRoad simulator.
    The open-loop plan shows a predicted trajectory with sampled inputs from the test set $\test$. For the open-loop behavior of BC, we only report the first control, marked with an orange cross. The position constraints are either the forward prediction (open-loop) or the driven trajectory of the leading vehicle (closed-loop). In the closed-loop control example, the dashed line jumps because of a cut-in, thus a different car becomes the leading vehicle.}
    \label{fig:open_close}
\end{figure}

To test the closed-loop performance, we benchmarked our learned models with the composable benchmark simulator CommonRoad
\cite{althoffCommonRoadComposableBenchmarks2017}
on real and synthetic scenarios.
The real scenarios were created from the HighD \cite{krajewskiHighDDatasetDrone2018} dataset, a collection of vehicle trajectories recorded on German highways, and then transformed into the CommonRoad format.
As the real scenarios do not cover the whole space of potential driving maneuvers we additionally created synthetic scenarios: (1) Breaking, involving uniformly sampled lead cars with decreasing acceleration, (2) Synthetic speed limit changes, which include uniformly sampled lead cars with either decreasing or increasing speed limit changes and (3) Synthetic cut-ins, where cut-ins in front of the ego car are generated with velocities uniformly sampled.
The trajectories of the lead car were generated via the intelligent driver model (IDM) \cite{treiberCongestedTrafficStates2000}.
The average scenario duration is around 6.5 seconds and we tested on $1540$ scenarios.
In \figref{fig:open_close} we can further see that the predicted open-loop trajectory takes into account constraints such as the terminal acceleration constraint $a_N = 0$.

\subsection{Experimental Setup}

\paragraph{Dataset} For solving the MPC formulation we used the NLP solver \texttt{acados} \cite{Verschueren2021}
with \texttt{HPIPM} \cite{Frison2020a} using a horizon length of $N=30$.
For creating training data, we uniformly sampled the initial state $\bar{x}_0$ from the box constraints of the longitudinal planner and filtered out scenarios where a crash was inevitable.
With a probability of $1/3$, we sampled a speed limit change and with a probability of $1/3$ we sampled that the lead car made a cut-in, leading to a different forward prediction.
We created a training dataset $\train$ with $10^5$ samples, a validation dataset $\valid$ with $33333$, samples and a test set $\test$ with $33333$ samples.

\paragraph{Metrics} We use the following metrics averaged over three training seeds: 
1) \emph{Trajectory MSE}: Is the mean squared error between the predicted trajectory and the ground truth $\hat{X}^\theta$ and $X^\textrm{MPC}$ on the test set $\test$.
2) \emph{Policy MSE}: Is the mean squared error between $\hat{u}_0^\theta$ and $u_0^\textrm{MPC}$ on the test set $\test$.
3) \emph{Planning Time}: We randomly sampled $1000$ inputs and estimate their inference times.
As there is some additive CPU noise e.g. from background processes, we repeated the inference benchmark $20 $ times for each input, taking the minimum.
Across all samples, we took the $95\%$ quantile to estimate the worst-case runtime.
4) \emph{Policy Time}: Similar to the planning time estimation, we estimated the inference time, but only for the first control.
5) \emph{Avg. $\lVert s \rVert_2$}, \emph{Avg. $\lVert v \rVert_2$} and \emph{Avg. $\lVert a \rVert_2$}: Are the average $\ell^2$-distances between the driven trajectories of the closed-loop MPC policy and the learned policy, in terms of the position $s$, velocity $v$, and acceleration $a$.

\paragraph{Hyperparameters} For each neural network architecture and loss combination, we did a hyperparameter search using the Optuna library \cite{akibaOptunaNextgenerationHyperparameter2019}.
The search space was defined by $[1\times 10^{-5}, 1\times 10^{-2}]$ for the learning rate, $\{1, 2, 3\}$ for the number of hidden layers in $\pi_\theta$, $\{32, 64, 128, 256, 512\}$ for the width of the hidden layers and $\{8, 16, 32, 64\}$ for the batch size.
For the policy part of the \texttt{PlanNetX} architectures and the baselines, we found that a width of $512$ and a depth of $3$ worked the best, while using \texttt{ReLU} activation functions.
Thus, this was also the model that we used for the pruning and quantization experiments.
For the \texttt{PlanNetXEnc} experiments we used three self-attention Transformer layers with embedding dimension $32$ and $4$ attention heads.
For the hyperparameter search, we used $100$ training epochs, whereas for the full training, we used $300$ epochs.
In all experiments, we used the Adam optimizer \cite{kingmaAdamMethodStochastic2015}.
For the weight matrix $W$ in \eqref{eq:uloss} and \eqref{eq:xloss}, we always used the identity matrix, as this gave us the best performance.
For the discount factor $\gamma$ we always used a value of $0.98$.
For measuring the inference time, we used an
AMD Ryzen 9 5950X @ 3.40 GHz CPU.

\paragraph{Neural Network Compression}
To reduce inference time further, we considered pruning and quantization:
Pruning removes the weights of the neural network that contribute little to the prediction.
We used iterative $\ell^1$-\emph{structured pruning} to prune the columns of the linear layers, \cite{liPruningFiltersEfficient2017}, which removes the columns with the smallest $\ell^1$-norm of one linear layer.
In our experiments, we set the number of parameters to be pruned to $90\%$.
Quantization is a technique that involves reducing the precision of the parameters of the network, typically from $32$-bit floating-point (FP32) to $16$-bit floating point (FP16) or $8$-bit integer (INT8) \cite{DBLP:journals/corr/abs-2303-17951}.
We used uniform quantization with INT8 precision, a common method with low overhead where the mapping consists of scaling, rounding and adding an offset. To estimate the clipping ranges and scaling factors, we employ post-training \emph{static quantization} using the validation dataset $\valid$ to collect statistics on the data before online inference \cite{gholamiSurveyQuantizationMethods2021}.

\subsection{Results}


\paragraph{Open-loop performance}

\begin{wraptable}{r}{0.4\textwidth}
    \centering
    \resizebox{\linewidth}{!}{%
    \begin{tabular}{lcc}
    \toprule
        Method & Policy MSE & Trajectory MSE \\
    \midrule
        BC & 0.66 & - \\
    \midrule
        \texttt{PlanNetX} & 107.68 & 0.01 \\
        \texttt{PlanNetXEnc} & 96.72 & 0.02 \\
        \texttt{PlanNetX} with $\mathcal{L}^\textrm{u}$ & 9.37 & 0.84  \\
    \bottomrule
    \end{tabular}
    }
    \caption{Open-loop performance}
    \label{tab:open_loop_results}
\end{wraptable}
For the open-loop planning performance, we measured the \emph{Policy MSE} and \emph{Trajectory MSE} on the test dataset $\test$.
From table~\ref{tab:open_loop_results}, we can see that the proposed methods \texttt{PlanNetX} and \texttt{PlanNetXEnc} can approximate the MPC plan policy very well especially when compared to the \texttt{PlanNetX} network with the $\mathcal{L}^u$ loss.
Interestingly, the \emph{Policy MSE} of the \texttt{PlanNetX} methods that use the state-trajectory loss is significantly higher than of BC, indicating that sometimes the \texttt{PlanNetX} deviates from the MPC plan to find its control sequence that leads to better tracking in state space.
We noticed that this was especially true when the MPC planner showed bang-bang behavior for the control sequence.

\paragraph{Closed-loop performance} When applying the approach \texttt{PlanNetX} in simulation for closed-loop control, we find that the driven trajectories are closer to the trajectories of the MPC policy than when just using the BC baseline, see \figref{fig:open_close} and table~\ref{tab:open_loop_results}.
The \texttt{PlanNetXEnc} achieves a similar performance then \texttt{PlanNetX}, whereas the control trajectory loss from \eqref{eq:uloss} leads to significantly worse results, indicating that learning the full control trajectory $U^\star$ does not necessarily help for learning the first control $u^\star_0$.
For the inference time of the control policy, we achieve a significant speed-up over the MPC solver.
Even though, for the planning time, the speed up is not as significant, we were able to reduce the planning time using structured $\ell^1$-pruning, without losing significantly in closed-loop performance.
The quantization with INT8 precision resulted in faster inference, but at the cost of a severe degradation in accuracy.
We suspect that during the roll-out of the \texttt{PlanNetX} architecture the rounding errors accumulate, leading to sub-optimal training results.
We therefore further tested quantization by using FP16, which maintains approximately the same accuracy with less memory consumption.
However, on a CPU we did not achieve any speedup.

\begin{table}[t]
    \centering
    \resizebox{\textwidth}{!}{%
    \begin{tabular}{lcccccc}
    \toprule
        Method & Avg. $\lVert \Delta s \rVert_2$ [m] & Avg. $\lVert \Delta v \rVert_2$ [m/s] & Avg. $\lVert \Delta a \rVert_2$ [m/$\textrm{s}^2$] & Policy Time [ms] & Planning Time [ms] \\
    \midrule
        MPC & - & - & - & 6.44 & 6.44 \\
        MPC warm start & - & - & - & 5.08 & 5.08 \\
        BC & 0.65 & 0.17 & 0.05 & 0.06 & - \\
    \midrule
        \texttt{PlanNetX} & 0.24 & 0.05 & 0.04 & 0.10 & 2.62 \\
        \texttt{PlanNetXEnc} & 0.29 & 0.05 & 0.03 & 0.31 & 2.79\\
        \texttt{PlanNetX} with $\mathcal{L}^u$ & 2.00 & 0.56 & 0.13 & 0.10 & 2.61 \\
        \texttt{PlanNetX} quant. & 1.19 & 0.39 & 0.19 & 0.07 & 1.94 \\
        \texttt{PlanNetX} $\ell^1$-str & 0.27 & 0.11 & 0.06 & 0.06 & 1.83 \\
    \bottomrule
    \end{tabular}
    }
    \caption{The closed-loop performance for the proposed \texttt{PlanNetX} and \texttt{PlanNetXEnc} approaches and some further ablations, where we tested the control trajectory loss $\mathcal{L}^u$, a pruned and INT8 quantized network as well as warm starting the MPC.}
    \label{tab:imitation_learning_results}
\end{table}

\paragraph{When to use the encoder?}
\begin{wrapfigure}{r}{0.45\textwidth}
    \centering
    \resizebox{\linewidth}{!}{%
    \includegraphics{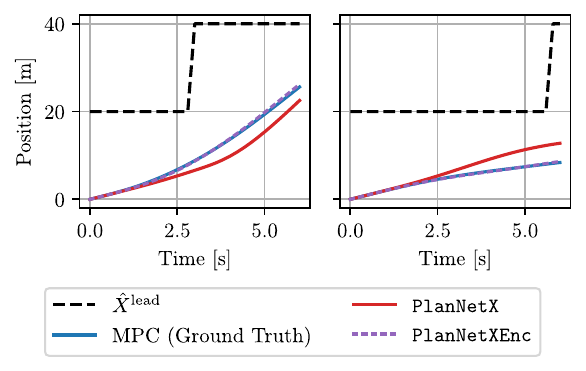}
    }
    \caption{Trajectory predictions for varying forward pred. $\hat{X}^\textrm{lead}$.}
    \label{fig:predictions}
\end{wrapfigure}
We further tested the capabilities of the encoder architecture to adapt to varying forward predictions of the lead car $\hat{X}^\textrm{lead}$.
For this, we introduced a jump in the forward prediction of the lead car at a randomized time point, c.f. \figref{fig:predictions}.
In such a scenario, the \texttt{PlanNetX} can not infer the information of when the jump occurs from only the stage-wise parameter $p_k$, thus taking an average guess to predict $\hat{x}_k$ and $\hat{u}_k$.
As we can see in \figref{fig:predictions}, by additionally using the encoder, the \texttt{PlanNetXEnc} is able to fit the MPC plan, as $z_k$ encodes the combined information of the parameters of all stages.
We defer a more detailed analysis of the generalization capabilities of the encoder for future research.

\section{Conclusion}
We found that our proposed \texttt{PlanNetX} framework can imitate the trajectories of the longitudinal MPC planner to high accuracy while achieving a better closed-loop control in simulation than BC and having a significantly lower inference time than the MPC planner.
Using compression techniques like pruning and quantization further reduced the inference time, however only with pruning we were able to obtain comparable performance to the uncompressed network.
Further, directions could be to extend the method to MPC with a non-linear model where stability issues due to vanishing or exploding gradients might arise.
Here, reducing the roll-out horizon, or using an MPC-based value function \cite{ghezziImitationLearningNonlinear2023} might further stabilize the training.
Other interesting directions would be to investigate the generalization capabilities of the \texttt{PlanNetXEnc} architecture for more MPC formulations with a potentially more complex parameterization of the MPC or to compare to other state-of-the-art IL methods like Diffusion models \cite{jannerPlanningDiffusionFlexible2022}.

\acks{
The authors thank Rudolf Reiter for providing helpful feedback throughout this work.
This project has been supported by the European Union’s Horizon 2020 research and innovation programme,
Marie Skłodowska-Curie grant agreement 953348, ELO-X, and by the Deutsche Forschungsgemeinschaft (DFG, German Research Foundation) with grant number 428605208.}

\bibliography{imitation}

\begin{thebibliography}{37}
\providecommand{\natexlab}[1]{#1}
\providecommand{\url}[1]{\texttt{#1}}
\expandafter\ifx\csname urlstyle\endcsname\relax
  \providecommand{\doi}[1]{doi: #1}\else
  \providecommand{\doi}{doi: \begingroup \urlstyle{rm}\Url}\fi

\bibitem[Akiba et~al.(2019)Akiba, Sano, Yanase, Ohta, and Koyama]{akibaOptunaNextgenerationHyperparameter2019}
Takuya Akiba, Shotaro Sano, Toshihiko Yanase, Takeru Ohta, and Masanori Koyama.
\newblock Optuna: {{A Next-generation Hyperparameter Optimization Framework}}.
\newblock In Ankur Teredesai, Vipin Kumar, Ying Li, R{\'o}mer Rosales, Evimaria Terzi, and George Karypis, editors, \emph{Proceedings of the 25th {{ACM SIGKDD International Conference}} on {{Knowledge Discovery}} \& {{Data Mining}}, {{KDD}} 2019, {{Anchorage}}, {{AK}}, {{USA}}, {{August}} 4-8, 2019}, pages 2623--2631. ACM, 2019.

\bibitem[Althoff et~al.(2017)Althoff, Koschi, and Manzinger]{althoffCommonRoadComposableBenchmarks2017}
Matthias Althoff, Markus Koschi, and Stefanie Manzinger.
\newblock {{CommonRoad}}: {{Composable}} benchmarks for motion planning on roads.
\newblock In \emph{2017 {{IEEE Intelligent Vehicles Symposium}} ({{IV}})}, pages 719--726, June 2017.

\bibitem[Bemporad et~al.(1999)Bemporad, Borrelli, and Morari]{Bemporad1999}
A.~Bemporad, F.~Borrelli, and M.~Morari.
\newblock The explicit solution of constrained {{LP-Based}} receding horizon control.
\newblock In \emph{Proceedings of the {{IEEE}} Conference on Decision and Control ({{CDC}})}, Sydney, Australia, 1999.

\bibitem[Carius et~al.(2020)Carius, Farshidian, and Hutter]{cariusMPCNetFirstPrinciples2020}
Jan Carius, Farbod Farshidian, and Marco Hutter.
\newblock {{MPC-Net}}: {{A First Principles Guided Policy Search}}.
\newblock \emph{IEEE Robotics and Automation Letters}, 5\penalty0 (2):\penalty0 2897--2904, April 2020.

\bibitem[Cosner et~al.(2022)Cosner, Yue, and Ames]{cosnerEndtoEndImitationLearning2022}
Ryan~K. Cosner, Yisong Yue, and Aaron~D. Ames.
\newblock End-to-{{End Imitation Learning}} with {{Safety Guarantees}} using {{Control Barrier Functions}}.
\newblock In \emph{2022 {{IEEE}} 61st {{Conference}} on {{Decision}} and {{Control}} ({{CDC}})}, pages 5316--5322, Cancun, Mexico, December 2022. IEEE.

\bibitem[Devlin et~al.(2019)Devlin, Chang, Lee, and Toutanova]{devlinBERTPretrainingDeep2019a}
Jacob Devlin, Ming-Wei Chang, Kenton Lee, and Kristina Toutanova.
\newblock {{BERT}}: {{Pre-training}} of {{Deep Bidirectional Transformers}} for {{Language Understanding}}.
\newblock In Jill Burstein, Christy Doran, and Thamar Solorio, editors, \emph{Proceedings of the 2019 {{Conference}} of the {{North American Chapter}} of the {{Association}} for {{Computational Linguistics}}: {{Human Language Technologies}}, {{NAACL-HLT}} 2019, {{Minneapolis}}, {{MN}}, {{USA}}, {{June}} 2-7, 2019, {{Volume}} 1 ({{Long}} and {{Short Papers}})}, pages 4171--4186. Association for Computational Linguistics, 2019.

\bibitem[Dosovitskiy et~al.(2021)Dosovitskiy, Beyer, Kolesnikov, Weissenborn, Zhai, Unterthiner, Dehghani, Minderer, Heigold, Gelly, Uszkoreit, and Houlsby]{dosovitskiyImageWorth16x162021}
Alexey Dosovitskiy, Lucas Beyer, Alexander Kolesnikov, Dirk Weissenborn, Xiaohua Zhai, Thomas Unterthiner, Mostafa Dehghani, Matthias Minderer, Georg Heigold, Sylvain Gelly, Jakob Uszkoreit, and Neil Houlsby.
\newblock An {{Image}} is {{Worth}} 16x16 {{Words}}: {{Transformers}} for {{Image Recognition}} at {{Scale}}.
\newblock In \emph{9th {{International Conference}} for {{Learning Representations}}}. International Conference for Learning Representations, 2021.

\bibitem[Drgo{\v n}a et~al.(2021)Drgo{\v n}a, Tuor, Skomski, Vasisht, and Vrabie]{drgonaDeepLearningExplicit2021}
J{\'a}n Drgo{\v n}a, Aaron Tuor, Elliott Skomski, Soumya Vasisht, and Draguna Vrabie.
\newblock Deep {{Learning Explicit Differentiable Predictive Control Laws}} for {{Buildings}}.
\newblock \emph{IFAC-PapersOnLine}, 54\penalty0 (6):\penalty0 14--19, January 2021.

\bibitem[Drgo{\v n}a et~al.(2024)Drgo{\v n}a, Tuor, and Vrabie]{drgonaLearningConstrainedParametric2024}
J{\'a}n Drgo{\v n}a, Aaron Tuor, and Draguna Vrabie.
\newblock Learning {{Constrained Parametric Differentiable Predictive Control Policies With Guarantees}}.
\newblock \emph{IEEE Transactions on Systems, Man, and Cybernetics: Systems}, pages 1--12, 2024.

\bibitem[Frison and Diehl(2020)]{Frison2020a}
G.~Frison and M.~Diehl.
\newblock {{HPIPM}}: A high-performance quadratic programming framework for model predictive control.
\newblock In \emph{Proceedings of the {{IFAC}} World Congress}, Berlin, Germany, July 2020.

\bibitem[Ghezzi et~al.(2023)Ghezzi, Hoffman, Frey, Boedecker, and Diehl]{ghezziImitationLearningNonlinear2023}
Andrea Ghezzi, Jasper Hoffman, Jonathan Frey, Joschka Boedecker, and Moritz Diehl.
\newblock Imitation {{Learning}} from {{Nonlinear MPC}} via the {{Exact Q-Loss}} and its {{Gauss-Newton Approximation}}.
\newblock In \emph{Proceedings of the {{IEEE Conference}} on {{Decision}} and {{Control}} ({{CDC}})}, 2023.

\bibitem[Gholami et~al.(2021)Gholami, Kim, Dong, Yao, Mahoney, and Keutzer]{gholamiSurveyQuantizationMethods2021}
Amir Gholami, Sehoon Kim, Zhen Dong, Zhewei Yao, Michael~W. Mahoney, and Kurt Keutzer.
\newblock A {{Survey}} of {{Quantization Methods}} for {{Efficient Neural Network Inference}}.
\newblock In \emph{Low-{{Power Computer Vision}}}, pages 291--326. {Chapman and Hall/CRC}, June 2021.

\bibitem[Gutjahr et~al.(2017)Gutjahr, Gr{\"o}ll, and Werling]{gutjahrLateralVehicleTrajectory2017}
Benjamin Gutjahr, Lutz Gr{\"o}ll, and Moritz Werling.
\newblock Lateral {{Vehicle Trajectory Optimization Using Constrained Linear Time-Varying MPC}}.
\newblock \emph{IEEE Transactions on Intelligent Transportation Systems}, 18\penalty0 (6):\penalty0 1586--1595, June 2017.

\bibitem[Hart et~al.(2019)Hart, Rychly, and Knoll]{hartLaneMergingUsingPolicybased2019}
Patrick Hart, Leonard Rychly, and Alois Knoll.
\newblock Lane-{{Merging Using Policy-based Reinforcement Learning}} and {{Post-Optimization}}.
\newblock In \emph{2019 {{IEEE Intelligent Transportation Systems Conference}} ({{ITSC}})}, pages 3176--3181, Auckland, New Zealand, October 2019. IEEE.

\bibitem[Hertneck et~al.(2018)Hertneck, Kohler, Trimpe, and Allgower]{hertneckLearningApproximateModel2018}
Michael Hertneck, Johannes Kohler, Sebastian Trimpe, and Frank Allgower.
\newblock Learning an {{Approximate Model Predictive Controller With Guarantees}}.
\newblock \emph{IEEE Control Systems Letters}, 2\penalty0 (3):\penalty0 543--548, July 2018.

\bibitem[Janner et~al.(2022)Janner, Du, Tenenbaum, and Levine]{jannerPlanningDiffusionFlexible2022}
Michael Janner, Yilun Du, Joshua Tenenbaum, and Sergey Levine.
\newblock Planning with {{Diffusion}} for {{Flexible Behavior Synthesis}}.
\newblock In \emph{International {{Conference}} on {{Machine Learning}}}, pages 9902--9915. PMLR, 2022.

\bibitem[Johansen and Grancharova(2003)]{Johansen2003}
T.A. Johansen and A.~Grancharova.
\newblock Approximate explicit constrained linear model predictive control via orthogonal search tree.
\newblock \emph{IEEE Trans. Automatic Control}, 48:\penalty0 810--815, 2003.

\bibitem[Karg et~al.(2021)Karg, Alamo, and Lucia]{kargProbabilisticPerformanceValidation2021}
Benjamin Karg, Teodoro Alamo, and Sergio Lucia.
\newblock Probabilistic performance validation of deep learning-based robust {{NMPC}} controllers.
\newblock \emph{International Journal of Robust and Nonlinear Control}, 31\penalty0 (18):\penalty0 8855--8876, 2021.

\bibitem[Kingma and Ba(2015)]{kingmaAdamMethodStochastic2015}
Diederik~P. Kingma and Jimmy Ba.
\newblock Adam: {{A Method}} for {{Stochastic Optimization}}.
\newblock In \emph{3rd {{International Conference}} for {{Learning Representations}}}, volume 3rd. International Conference for Learning Representations, 2015.

\bibitem[Kiran et~al.(2022)Kiran, Sobh, Talpaert, Mannion, Sallab, Yogamani, and P{\'e}rez]{kiranDeepReinforcementLearning2022b}
B~Ravi Kiran, Ibrahim Sobh, Victor Talpaert, Patrick Mannion, Ahmad A.~Al Sallab, Senthil Yogamani, and Patrick P{\'e}rez.
\newblock Deep {{Reinforcement Learning}} for {{Autonomous Driving}}: {{A Survey}}.
\newblock \emph{IEEE Transactions on Intelligent Transportation Systems}, 23\penalty0 (6):\penalty0 4909--4926, June 2022.

\bibitem[Krajewski et~al.(2018)Krajewski, Bock, Kloeker, and Eckstein]{krajewskiHighDDatasetDrone2018}
Robert Krajewski, Julian Bock, Laurent Kloeker, and Lutz Eckstein.
\newblock The {{highD Dataset}}: {{A Drone Dataset}} of {{Naturalistic Vehicle Trajectories}} on {{German Highways}} for {{Validation}} of {{Highly Automated Driving Systems}}.
\newblock In \emph{2018 21st {{International Conference}} on {{Intelligent Transportation Systems}} ({{ITSC}})}, pages 2118--2125, 2018.

\bibitem[Levine and Koltun(2013{\natexlab{a}})]{levineGuidedPolicySearch2013}
Sergey Levine and Vladlen Koltun.
\newblock Guided {{Policy Search}}.
\newblock In \emph{Proceedings of the 30th {{International Conference}} on {{Machine Learning}}}, pages 1--9. PMLR, May 2013{\natexlab{a}}.

\bibitem[Levine and Koltun(2013{\natexlab{b}})]{levineVariationalPolicySearch2013}
Sergey Levine and Vladlen Koltun.
\newblock Variational {{Policy Search}} via {{Trajectory Optimization}}.
\newblock In \emph{Advances in {{Neural Information Processing Systems}}}, volume~26, 2013{\natexlab{b}}.

\bibitem[Levine et~al.(2016)Levine, Finn, Darrell, and Abbeel]{levineEndtoendTrainingDeep2016}
Sergey Levine, Chelsea Finn, Trevor Darrell, and Pieter Abbeel.
\newblock End-to-end training of deep visuomotor policies.
\newblock \emph{Journal of Machine Learning Research}, 17\penalty0 (39):\penalty0 1--40, 2016.

\bibitem[Li et~al.(2017)Li, Kadav, Durdanovic, Samet, and Graf]{liPruningFiltersEfficient2017}
Hao Li, Asim Kadav, Igor Durdanovic, Hanan Samet, and Hans~Peter Graf.
\newblock Pruning {{Filters}} for {{Efficient ConvNets}}.
\newblock In \emph{5th {{International Conference}} on {{Learning Representations}}}. International Conference on Learning Representations, 2017.

\bibitem[Li and Todorov(2004)]{li2004iterative}
Weiwei Li and Emanuel Todorov.
\newblock Iterative linear quadratic regulator design for nonlinear biological movement systems.
\newblock In \emph{First International Conference on Informatics in Control, Automation and Robotics}, volume~2, pages 222--229. SciTePress, 2004.

\bibitem[Mamedov et~al.(2022)Mamedov, Reiter, Diehl, and Swevers]{mamedovSafeImitationLearning2022}
Shamil Mamedov, Rudolf Reiter, Moritz Diehl, and Jan Swevers.
\newblock Safe {{Imitation Learning}} of {{Nonlinear Model Predictive Control}} for {{Flexible Robots}}.
\newblock \emph{CoRR}, abs/2212.02941, 2022.

\bibitem[Mordatch and Todorov(2014)]{mordatchCombiningBenefitsFunction2014}
Igor Mordatch and Emo Todorov.
\newblock Combining the benefits of function approximation and trajectory optimization.
\newblock In \emph{Robotics: {{Science}} and {{Systems X}}}. {Robotics: Science and Systems Foundation}, July 2014.

\bibitem[Pek and Althoff(2021)]{pekFailSafeMotionPlanning2021}
Christian Pek and Matthias Althoff.
\newblock Fail-{{Safe Motion Planning}} for {{Online Verification}} of {{Autonomous Vehicles Using Convex Optimization}}.
\newblock \emph{IEEE Transactions on Robotics}, 37\penalty0 (3):\penalty0 798--814, June 2021.

\bibitem[Reske et~al.(2021)Reske, Carius, Ma, Farshidian, and Hutter]{reskeImitationLearningMPC2021}
Alexander Reske, Jan Carius, Yuntao Ma, Farbod Farshidian, and Marco Hutter.
\newblock Imitation {{Learning}} from {{MPC}} for {{Quadrupedal Multi-Gait Control}}.
\newblock In \emph{2021 {{IEEE International Conference}} on {{Robotics}} and {{Automation}} ({{ICRA}})}, pages 5014--5020, May 2021.

\bibitem[Schwan et~al.(2023)Schwan, Jones, and Kuhn]{schwanStabilityVerificationNeural2023}
Roland Schwan, Colin~N. Jones, and Daniel Kuhn.
\newblock Stability {{Verification}} of {{Neural Network Controllers Using Mixed-Integer Programming}}.
\newblock \emph{IEEE Transactions on Automatic Control}, pages 1--16, 2023.

\bibitem[Treiber et~al.(2000)Treiber, Hennecke, and Helbing]{treiberCongestedTrafficStates2000}
Martin Treiber, Ansgar Hennecke, and Dirk Helbing.
\newblock Congested {{Traffic States}} in {{Empirical Observations}} and {{Microscopic Simulations}}.
\newblock \emph{Physical Review E}, 62\penalty0 (2):\penalty0 1805--1824, August 2000.

\bibitem[van Baalen et~al.(2023)van Baalen, Kuzmin, Nair, Ren, Mahurin, Patel, Subramanian, Lee, Nagel, Soriaga, and Blankevoort]{DBLP:journals/corr/abs-2303-17951}
Mart van Baalen, Andrey Kuzmin, Suparna~S. Nair, Yuwei Ren, Eric Mahurin, Chirag Patel, Sundar Subramanian, Sanghyuk Lee, Markus Nagel, Joseph Soriaga, and Tijmen Blankevoort.
\newblock {FP8} versus {INT8} for efficient deep learning inference.
\newblock \emph{CoRR}, abs/2303.17951, 2023.

\bibitem[Vaswani et~al.(2017)Vaswani, Shazeer, Parmar, Uszkoreit, Jones, Gomez, Kaiser, and Polosukhin]{vaswaniAttentionAllYou2017a}
Ashish Vaswani, Noam Shazeer, Niki Parmar, Jakob Uszkoreit, Llion Jones, Aidan~N Gomez, {\L}ukasz Kaiser, and Illia Polosukhin.
\newblock Attention is {{All}} you {{Need}}.
\newblock In \emph{Advances in {{Neural Information Processing Systems}}}, volume~30, 2017.

\bibitem[Vaupel et~al.(2020)Vaupel, Hamacher, Caspari, Mhamdi, Kevrekidis, and Mitsos]{vaupelAcceleratingNonlinearModel2020}
Yannic Vaupel, Nils~C. Hamacher, Adrian Caspari, Adel Mhamdi, Ioannis~G. Kevrekidis, and Alexander Mitsos.
\newblock Accelerating nonlinear model predictive control through machine learning.
\newblock \emph{Journal of Process Control}, 92:\penalty0 261--270, August 2020.

\bibitem[Verschueren et~al.(2021)Verschueren, Frison, Kouzoupis, Frey, {van Duijkeren}, Zanelli, Novoselnik, Albin, Quirynen, and Diehl]{Verschueren2021}
Robin Verschueren, Gianluca Frison, Dimitris Kouzoupis, Jonathan Frey, Niels {van Duijkeren}, Andrea Zanelli, Branimir Novoselnik, Thivaharan Albin, Rien Quirynen, and Moritz Diehl.
\newblock Acados -- a modular open-source framework for fast embedded optimal control.
\newblock \emph{Mathematical Programming Computation}, October 2021.

\bibitem[Wabersich and Zeilinger(2021)]{wabersichPredictiveSafetyFilter2021}
Kim~Peter Wabersich and Melanie~N. Zeilinger.
\newblock A predictive safety filter for learning-based control of constrained nonlinear dynamical systems.
\newblock \emph{Automatica}, 129, July 2021.

\end{thebibliography}

\end{document}